\documentclass[10pt,journal,compsoc]{IEEEtran}

\usepackage[nocompress]{cite}
\usepackage{graphics}
\usepackage{epsfig}
\usepackage[colorlinks]{hyperref}
\usepackage{amssymb,amsmath,mathptmx,amsbsy,bm}

\usepackage{adjustbox}
\usepackage{boites,boites_examples,graphicx}

\begin{document}

\title{Single Image Depth Estimation: \\An Overview}

\author{Alican~Mertan,
        Damien~Jade~Duff,
        and~Gozde~Unal
\IEEEcompsocitemizethanks{\IEEEcompsocthanksitem Istanbul Technical University, Istanbul, Turkey

\IEEEcompsocthanksitem Corresponding author's e-mail: mertana@itu.edu.tr.}% <-this % stops an unwanted space
\thanks{This work is based on Alican Mertan's Master's thesis \cite{mertan_application_2020}}}

\IEEEtitleabstractindextext{%
\begin{abstract}
We review solutions to the problem of depth estimation, arguably the most important subtask in scene understanding. We focus on the single image depth estimation problem.  Due to its properties, the single image depth estimation problem is currently best tackled with machine learning methods, most successfully with convolutional neural networks. We provide an overview of the field by examining key works. We examine non-deep learning approaches that mostly predate deep learning and utilize hand-crafted features and assumptions, and more recent works that mostly use deep learning techniques. The single image depth estimation problem is tackled first in a supervised fashion with absolute or relative depth information acquired from human or sensor-labeled data, or in an unsupervised way using unlabelled stereo images or video datasets. We also study multitask approaches that combine the depth estimation problem with related tasks such as semantic segmentation and surface normal estimation. Finally, we discuss investigations into the mechanisms, principles and failure cases of contemporary solutions.
\end{abstract}

}

\maketitle

\IEEEraisesectionheading{\section{Introduction}\label{sec:introduction}}

\begin{figure}[h]
\centering
\resizebox{\columnwidth}{!}{
\begin{tabular}{ cc } 
 \includegraphics[width=1.4in,height=0.7in]{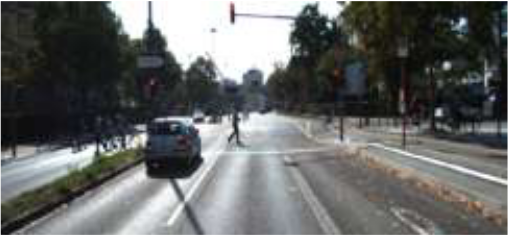}
 & 
 \includegraphics[width=1.4in,height=0.7in]{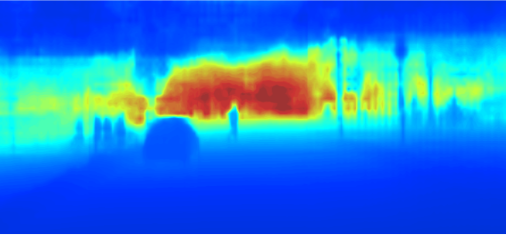}
 \\ 
\end{tabular}}
\caption{ Input RGB image and the depth map estimated by the neural network of Fu et al. \cite{fu_deep_2018}. }
\label{fig:exampleSIDE}
\end{figure}

\IEEEPARstart{D}{epth} estimation from a single image (SIDE, short for Single Image Depth Estimation) is the task of estimating a dense depth map for a given single RGB image. More specifically, for each pixel in the given RGB image, one needs to estimate a metric depth value. An example of an input image and corresponding depth map can be seen in \textbf{Figure \ref{fig:exampleSIDE}.} Here, the colors in the depth map correspond to the depth of that pixel: blueish means the pixel is closer to us, reddish means the pixel is further away from us.

What makes the SIDE problem interesting and challenging is its inherent ambiguity. An endless number of different 3D scenes can result in the same 2D image. This suggests that there is a one to many mapping from RGB images to depth maps. If this is the case, how do human beings, whose visual systems highly surpass artificially created visual systems in terms of quality and generalization, estimate the depth from monocular images? The answer to this question lies in the cues humans use to do SIDE.

For estimating the depth from a single image, the human visual system is the most superior system in terms of quality and generalization. Foley and Maitlin \cite{foley_sensation_2015} catalog the known pictorial (static) monocular cues used by human beings to estimate depth from a single image. There are seven such static cues that we can use to estimate the depth from a static single image. The first cue is occlusion which happens when one object partially covers another one. The partially covered object is considered to be farther away. The second cue is called perspective. We can observe this by looking at parallel lines that appear to meet in the distance. Two other cues are related to the perspective. One of them is size cue. The same object can have different sizes on the retinal image according to its distance. Therefore, the size of an object influences our depth estimates. The second cue related to the perspective is texture gradient. It happens when you look at a surface at a slant. The texture of the surface becomes denser as the distance increases. Another cue that we use to infer depth is called the atmospheric cue. It refers to the observation that objects get blurry and bluish as they move away from us. Moreover, we use patterns of light and shadows when perceiving depth. We consider things like objects casting shadows onto other objects or having shadows attached to their surfaces. The last cue that we use is the height cue. Objects closer to the horizon seem farther away.

\begin{figure}
\centering
\resizebox{\columnwidth}{!}{
\begin{tabular}{ c } 
\includegraphics[width=2.5in,height=0.7in]{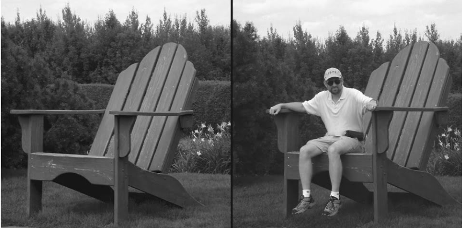}
\end{tabular}}
\caption{ An example of illusions that fools the human visual system. Image taken from \cite{foley_sensation_2015}. }
\label{fig:chairIllusion}
\end{figure}

The most important cue here is the size cue. As humans, we have a rough estimate of the objects' size that we see in real world everyday. When we look at the world and observe 2D RGB images, our visual system estimates the 3D scene between an endless number of geometrically possible 3D scenes, using our prior knowledge to choose the one that fits into the world as we know it. This is also the reason why we are fooled by the images that are similar to the ones in the \textbf{Figure \ref{fig:chairIllusion}.} Since there is no other cue that tells us otherwise, we assume the chair to have a usual size and accordingly estimate its depth closer to us. However, by looking at the relative sizes of the human and the chair in the right image, we understand that the chair is farther away than we estimated since it is bigger than that we assumed.

All this information leads us to a very important conclusion: we, as humans, use learned prior knowledge, and our visual system tends to work statistically \cite{hoiem_automatic_2005}. This conclusion also directs the way the research in this area is conducted and statistical methods are heavily utilized to solve the SIDE problem.

In this work, we offer an extensive overview of the learning-based solutions for the SIDE problem, in which we outline the research categories. Works in the categories are summarized in a way that highlights the logical progression of the solutions. Common themes that are seen in multiple works, problem-specific approaches, and insights are emphasized. 

\begin{figure*}
    \centering
    \includegraphics[width=\linewidth]{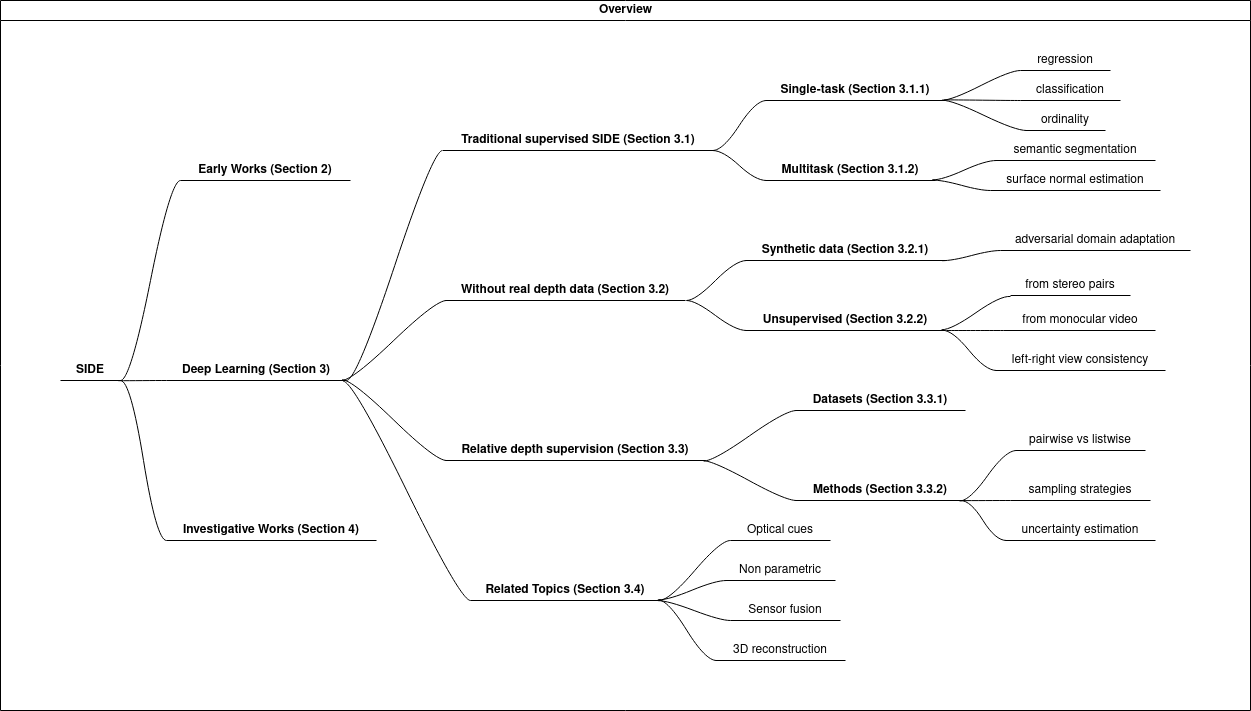}
    \caption{An overview of the SIDE problem. Bold font indicates sections, normal fonts indicate keywords about the methods presented in the corresponding section.}
    \label{fig:overview}
\end{figure*}

\textbf{Figure \ref{fig:overview}} shows the structure we use in our work to discuss the works.

\section{Early Days}

In the early days of the field, the SIDE problem was not tackled directly. In the classical work of Hoeim et al. \cite{hoiem_automatic_2005}, the authors aim to automatically reconstruct a 3D scene from a given RGB image for virtual environment creation. Their approach makes the assumption that outdoor environments basically consist of the sky, ground plane, and vertical objects sticking out of the ground. They use hand-generated cues to classify superpixels in one of the three classes. Afterward, using the three classes and the above-mentioned assumption, they automatically create the virtual environment by placing the objects on the ground plane vertically. Since the elements of the inferred scene are very simplified like a photo pop up from a child's book, there are some details missing. Nonetheless, the end results look pleasing to the eye. In this work, we see the first examples of two important recurring themes in the field: 
\begin{boitecoloriee}
\textbf{Incorporating semantic segmentation} Semantic information is incredibly important for estimation of depth. It can help a computer vision system to exploit its prior knowledge for a given semantic class. For example, by looking at two pieces of blue patches from an image, we can estimate their depth by knowing one of them is the sky and the other one is water. Even though semantic information is expected to be exploited by machine learning techniques implicitly, it also has been explicitly utilized to solve the SIDE problem by many researchers in different ways \cite{hoiem_automatic_2005, liu_single_2010, karsch_depth_2014, ladicky_pulling_2014, eigen_predicting_2015, wang_towards_2015}.

\textbf{Separating indoor and outdoor} Although humans seamlessly estimate the depth of indoor and outdoor scenes without noticing any change between them, they are actually structured very differently. 
For instance, the aforementioned assumptions made by Hoeim et al. \cite{hoiem_automatic_2005} do not hold for indoor scenes. Even when researchers make no assumption about the structure of the environment, they still let their system work on a single type of environment most of the time. This is because of the inherent difference between indoor and outdoor environments which makes the statistical learning of a system that works on both types of environments a challenging problem. Nevertheless, some of the work in the field actually tackled the problem of SIDE in the wild \cite{saxena_learning_2006, liu_discrete-continuous_2014, chen_single-image_2016}. Unless otherwise stated, all the works that are going to be mentioned are designed to work in either indoor or outdoor environments.
\end{boitecoloriee}

Another early work that uses SIDE to solve a given problem 
is the work of Michels et al. \cite{michels_high_2005}. The task in this work is to navigate a high-speed remote control car through obstacles in an uncontrolled outdoor environment. The designed framework consists of two parts: a vision part that fakes a 2D laser scanner and estimates the distance of the nearest obstacle in each direction, and a reinforcement learning part that drives the car avoiding obstacles. The reason why depth is estimated from a single image, in this case, is that it gives a better range compared to usually preferred binocular vision. The vision system is trained using linear regression with handcrafted features in a supervised manner. The input image is divided into vertical strips. Each strip is labeled with the nearest obstacle's distance in log space. Handcrafted features are created for each stripe while preserving their spatial information. In order to be able to capture the global context, the system uses neighboring stripes' features as well as the features of the stripe that it is making its decision for. Additionally, we see different error metrics being used. The distance between the estimated depth and the ground truth depth in log space is one of them. Also, relative depth error, where the mean is subtracted from estimated and ground-truth depth values in log space, is used as an error metric. Moreover, synthetic data with different levels of realism is used to boost the success of the system. Since the task here is to avoid obstacles, the vision problem is formulated very differently (the distance of the nearest obstacle in each direction) compared to our definition of SIDE (pixel by pixel, dense depth estimation). Yet we see very important ideas that have been repeated by the later works in the field:
\begin{boitecoloriee}
    \textbf{Working in log space} The challenge of estimating the depth of close objects and distant objects is not the same. While being a few centimeters off in our estimation of depth for an object that is meters away is acceptable, it is definitely a bigger mistake to be a few centimeters off if the object is only ten centimeters away. This is as much the case for humans as it is for computer vision systems. While we can be more precise in our estimations for smaller depths, we could only provide a rough depth range for bigger depths. For this reason, the errors are usually calculated in log space since the logarithm function maps the depth values in a way that error functions become more forgiving for mistakes in bigger depths. 
    
    \textbf{Using spatial coordinates} Spatial coordinates can be an important cue to estimate depth. It can help to exploit the structure of the scenes that the system has seen before. When working on a dataset consisting of outdoor images, the pixels in the upper rows have a high chance of being part of a sky, and statistical learning systems can exploit this kind of relations easily. On the other hand, this exploitation can create bias in the system.
    
    \textbf{Incorporating global context} As stated earlier, the scale of the scene is ambiguous. However, this ambiguity can be ignored in real-world scenarios since objects with known sizes can provide us enough cues to estimate the scale of the scene. These cues \iffalse are called global cues that\fi require us to look at a bigger part of the input image than just a local patch.
    
    \textbf{Using relative depth} The term "relative depth" refers to the ordering of the depth of pixels instead of absolute depth which refers to the metric depth values. Relative depth can be used to measure the performance of the system or can be used as an error function. This way, the system would not be penalized for mistakes due to scale ambiguity.
    
    \textbf{Training with synthetic data} Machine learning systems need lots of data points in order to be able to learn the task at hand. For the SIDE task, datasets consist of RGB images and corresponding depth maps. Unfortunately, acquiring RGB images and corresponding depth maps is a costly job. Even though different datasets have been collected throughout the last decades \cite{silberman_indoor_2012, geiger_vision_2013}, the need for labeled data is considered to be a problem in general in machine learning. One of the ways to overcome this problem is the usage of synthetic data. The system still learns with labeled data, however, synthetic data can be created and labeled automatically with ease in great amounts. Moreover, similar to data augmentation, great diversity in the data can be achieved by changing the texture of the objects or the lighting of the scene while keeping the ground truth the same, which will increase the robustness of the system. For all these reasons, the usage of synthetic data is considered to be a solution. On the other hand, the usage of synthetic data, in itself, creates a problem. Synthetic and real data are considered as different domains and the system needs to adapt to the real data after it is trained on synthetic data. Additionally, while diversity for a given scene can be achieved with ease, creating diverse sets of natural scenes synthetically is an incredibly time-consuming job.

\end{boitecoloriee}

To the best of our knowledge, the first work that tries to estimate a full metric depth map from a given single RGB image is the work of Saxena et al. \cite{saxena_learning_2006}. They process an input image in small patches by applying handcrafted filters to extract image features. For each small patch, a single depth value is estimated. To be able to successfully determine the absolute depth, global cues are incorporated by applying the aforementioned filters in multiple scales, utilizing the neighboring patches' features, and utilizing features from the same column. Features from the same image column are used based on the observation that most of the structures in the images are vertical. Additionally, to increase the understanding of the system for neighboring patches, histograms of the features are calculated for each patch and the system is fed with the difference between the histogram of a patch and its neighboring patches.

A Markov Random Field (MRF) model is trained in a supervised manner to estimate the metric depth from these features. Three different sets of parameters are learned from the training set for each row in the image. The reason for learning different parameters for different rows of the image is the observation that each row is statistically different from each other. The first set of learned parameters is for estimating absolute depth for a given patch by looking at the features. Another set of parameters tries to estimate the uncertainty in the absolute depth estimation. The last set of parameters are related to the smoothing term in the model. Neighboring patches usually have a very similar depth when they are similar in the RGB domain. Therefore a smoothing term is added to the model to bring neighboring patches' depth closer to each other. The effect of this term is controlled by the last set of parameters which determines the amount of the depth similarity of the patches. New themes encountered here are:
\begin{boitecoloriee}
    \textbf{Working at multiple scales} The same object can look very different when being looked at different distances. While we do not have images of the same scene taken from different distances for the SIDE task, we could simulate a similar situation by changing the resolution of the image. By working in multiple scales, the system becomes more flexible since it can look at different cues at different scales to detect the depth of objects.
    
    \textbf{Incorporating human knowledge} Incorporating prior human knowledge in terms of hand-designed filters, assumptions, and loss functions in MRF based models was a common practice. With the rise of deep learning techniques, this trend started to decline, yet we still see incorporated human knowledge in the form of network architecture design and loss functions.
    
    \textbf{Adding a smoothing term} A great example of added prior knowledge is smoothing term. We know that depth discontinuities only occur at the edge of objects. Depth maps mostly consist of smooth depth transitions where the neighboring pixels have a very similar depth. A smoothing term can be used in a loss function to achieve smoother depth maps. However, over smoothing of the actual depth discontinuities should be prevented.
\end{boitecoloriee}

In 2008, Saxena et al. published another work \cite{saxena_make3d:_2008} that influenced the field with a very substantial assumption: scenes consist of small planar surfaces and the depth of all the pixels belonging to a surface can be calculated by the 3D location and orientation of the surface they belong to. This basically means that even the most complex 3D scenes can be expressed with the 3D location and the orientation of small surfaces. The validity of this assumption can be seen in graphics engines where many complex models can be created with simple triangular surfaces. They create these small surfaces by superpixelating the image in the RGB domain with the expectation of similar-looking neighboring pixels belonging to the same surface. 

We see lots of similarities with their previous work \cite{saxena_learning_2006}, the biggest difference being the use of superpixels. In this work, they again use the MRF model and train it in a supervised manner. 

Image features are calculated via hand-designed filters. To incorporate global information, the features of the neighboring superpixels are used. Similar to their previous work \cite{saxena_learning_2006}, the MRF model estimates different parameters for different rows of the image to model the relationship between image features and the 3D location and orientation of the superpixels. In their model, they also try to capture three more characteristics from the image:
\renewcommand{\labelitemi}{\textendash}
\begin{itemize}
    \item superpixels that are connected in 3D, since most of the neighboring superpixels should be connected except for the case of occlusion;
    \item superpixels that are on the same plane in 3D, since most of the superpixels are not only just connected but also parts of the same plane if no edges can be found between them;
    \item straight lines in RGB domain as they are most likely to be straight lines in 3D.
\end{itemize}
It is important to capture such characteristics as more constraints can be added for depth estimation based on these characteristics. Fractional (relative) depth error is used while applying these constraints. It is formulated as $ \frac{( \hat{d} - d)}{d} $ where $ \hat{d} $ is estimated depth and the $ d $ is the ground truth depth value.

They also extend their work by detecting objects and using prior knowledge to better estimate the depth of detected object such as detecting a human and expecting it to be connected to the ground or detecting two humans and using their sizes in pixels to better estimate their depth (the one with twice a size of other in pixels is most likely to be closer to the camera). 

In this work, we see a very influential assumption that allows the SIDE problem to be simplified:
\renewcommand{\labelitemi}{\textbullet}
\begin{boitecoloriee}

    \textbf{Superpixelating the input image} Based on the assumption that a 3D scene can be expressed with small planar surfaces, many scientists superpixelate the input image and estimate the depth of the superpixels \cite{saxena_make3d:_2008, liu_single_2010, liu_discrete-continuous_2014, liu_single-view_2014, li_depth_2015, liu_deep_2015, wang_towards_2015, zoran_learning_2015, yan_single_2017}. This has the benefit of reducing the computational cost as the number of points for estimation is decreased by this assumption.
    
    \textbf{Coplanarity assumption} Superpixels that are similar in RGB domain are more likely to be on the same surface. This assumption is usually utilized to smooth the estimation of the model.
\end{boitecoloriee}

\section{SIDE Landscape}

We start by summarizing some of the datasets that are frequently used in the literature. For a better review of available datasets, please refer to \cite{firman_rgbd_2016}.

\textbf{NYU}\cite{silberman_indoor_2012} It consists of indoor images, and their corresponding depth images and semantic segmentation maps. It contains 1449 densely labeled images from 464 indoor scenes such as bathrooms, kitchens, libraries, living rooms, etc. Additionally, it has a raw version consists of over 400k images and corresponding depth maps coming from the video recording of the same indoor environments.

\textbf{KITTI}\cite{geiger_vision_2013} It consists of over 93k images with their corresponding depth maps, collected by driving a car in a mid-size city, in rural areas, and in highways. It contains outdoor images with visible cars and pedestrians. 

\textbf{Sun RGB-D}\cite{song_sun_2015} It contains 10,335 indoor images with corresponding depth maps. Additionally, it contains ground truth annotations for scene category, 2D and 3D bounding boxes for object detection, 2D semantic segmentation, object orientation, and room layout estimation.

\textbf{SceneNet RGB-D}\cite{mccormac_scenenet_2016} It contains synthetically created 5M indoor images with their depth maps. Room conditions, lighting, and textures are randomly generated. 

We group the works based on their main consideration. Each work in the field approaches the SIDE problem from a different perspective, trying to improve different aspects of the existing solutions. While some of the works try to achieve better metric results in the datasets by simply applying new developments in the techniques or utilizing multi-tasking, others try to overcome the need for labeled data, aim for better 3D structure, and so on. We group the works under 3 main categories as follows:
\begin{itemize}
    \item \textbf{Increasing the Metric Performance} Here we summarize the works that mainly focus on increasing the metric performance. We organize works under two groups. First, we present works that try to achieve better performance by architectural choices, loss functions, etc. Next, we present works that focus on multitask learning.
    \item \textbf{Eliminating the Need for Labeled Data} In this group, we present the works that solve the SIDE problem without annotated real-world ground truth. Here we see two main approaches, usage of synthetic data where the costly ground truth data collection is replaced by an automated creation of synthetic data, and unsupervised learning where geometry is utilized to acquire supervision. 
    \item \textbf{Working in the Wild} Lastly, we summarize the works that focus on SIDE in the wild. Since the SIDE problem is reformulated in these works, we present the available datasets and ground truth annotations first, then summarize the works on SIDE in the wild.
\end{itemize}{}
    \newcommand{\centered}[1]{\begin{tabular}{@{}l@{}} #1 \end{tabular}}
    \subsection{Main consideration: increasing the metric performance}
    In this section, we examine the works that are mainly focused on getting better metric results. Naturally, some works closely follow previous works and just apply new extensions and developments in the given techniques. On the other hand, there are works that incorporate human knowledge to boost performance. We also examine works that learn multiple tasks jointly to create synergy in order to get better results. 
    
        \subsubsection{CNN based works}
        
        Here we list the works that use CNNs at the core of their frameworks. Note that this will not be an exhaustive list of works that use CNN, instead we are going to list works that use CNN while focusing on getting better results through architectural choices, loss functions, and so on. \textbf{Table~\ref{tab:cnn}} summarizes the works that we discuss in this section.

\begin{table*}[]
\centering
\caption{\label{tab:cnn}Summary of the main works we discussed that solve the SIDE problem in a supervised single-task manner.}
\begin{adjustbox}{width=1\textwidth}
\begin{tabular}{llll}
 & \begin{tabular}{c}formulation\end{tabular} & \begin{tabular}{c}loss function\end{tabular} & \begin{tabular}{c}architecture summary\end{tabular}  \\ \hline\hline
 
 \centered{Eigen~et~al.~\cite{eigen_depth_2014}} & \centered{regression} & \centered{scale-invariant loss} & \centered{A convolutional network with fully connected layers outputs coarse \\estimation and a fully convolutional network refines the coarse estimations.} \\\hline
 
 \centered{Eigen~et~al.~\cite{eigen_predicting_2015}} & \centered{regression} & \centered{scale-invariant loss + \\gradient matching regularization \\term} & \centered{An additional stack of fully convolutional neural network is added to the \\architecture of \cite{eigen_depth_2014}, to further refine the predictions and increase the spatial \\resolution. } \\\hline
 
 \centered{Laina~et~al.~\cite{laina_deeper_2016}} & \centered{regression} & \centered{berHu loss} & \centered{It consists of a ResNet-50 based encoder and a decoder with up-convolutional \\blocks to increase the spatial resolution of the output.} \\\hline
 
 \centered{Cao~et~al.~\cite{cao_estimating_2017}} & \centered{classification} & \centered{cross-entropy loss weighted\\by an information gain matrix} & \centered{A ResNet based fully convolutional network outputs initial class predictions \\and a fully connected CRF refines the initial estimations.} \\\hline
 
 \centered{Fu~et~al.~\cite{fu_deep_2018}} & \centered{ordinal regression} & \centered{ordinal loss} & \centered{ It consists of a generic feature extractor part followed by a multi-scale feature \\learner which applies dilated convolutions, 1x1 convolutions to detect cross \\channel features, and a full image encoder to capture global information. These \\information then passed to the multiple heads of the network. }  \\\hline
 
 \centered{Bhat~et~al.~\cite{bhat2020adabins}} & \centered{hybrid regression} & \centered{scaled scale-invariant loss + \\bi-directional Chamfer loss as a \\regularization} & \centered{An EfficientNet based encoder and a standard decoder block outputs multi-\\channel feature maps which is used by a visual transformer to estimate bin \\widths and probabilities.} \\\hline
 
\end{tabular}
\end{adjustbox}
\end{table*}
        
In 2014, Eigen et al. \cite{eigen_depth_2014} introduce CNNs to the SIDE problem and achieve relatively good performance, when compared to earlier methods. CNN based solutions were already achieving quite satisfactory results in different vision problems at that time. Eigen et al. utilize the experience gained so far on CNNs and combine it with problem-specific knowledge to tackle the SIDE problem. They formulate the problem as a supervised regression problem and solve it with their framework, consisting of two networks, namely a $ Coarse $ and a $ Fine $ network, which are stacked on top of each other. Its operations can be summarized as follows:
\begin{eqnarray}
    Coarse(Input) &=& Depth_{coarse}\\
    Fine(Input, Depth_{coarse}) &=& Depth_{fine}
\end{eqnarray}{}
The $ Coarse $  network consists of convolutional layers and fully connected layers at the end. Because of the fully connected layers, we could say that the network makes its decision by looking at the image as a whole. This allows it to utilize the "global context" of the image and make a coarse estimation of the depth of a scene. However, fully connected layers come with a huge computational cost. In order to be able to keep the model reasonable in terms of memory, the output resolution is decreased.
        
The $ Fine $  network is a fully convolutional network and works by considering only the local parts of the image. It takes the original input image and the estimation of the coarse network. In a sense, it refines the coarse estimation by working locally. To deal with the scale ambiguity, the authors define a scale-invariant loss in the log space. Three reformulations of the same loss function can be seen below. $ \hat{d} $ and $ d $ are the predicted and the ground truth depth, respectively. Sub indices indicate the pixels, and $ n $ is the total number of pixels in an image.
\begin{equation}
    \label{eqn:eigenloss1}
    loss = \frac{1}{2n}\sum_{i=1}^n ( \log {\hat{d}_i} - \log d_i + \frac{1}{n} \sum_i ( \log d_i - \log \hat{d}_i ) )^2    
\end{equation}{}
\begin{equation}
    \label{eqn:eigenloss2}
    = \frac{1}{2n^2} \sum_{i,j} ( (\log {\hat{d}_i} - \log {\hat{d}_j}) - (\log {d_i} - \log {d_j}) )^2
\end{equation}{}
\begin{equation}
    \label{eqn:eigenloss3}
    = \frac{1}{n} \sum_i (\log {\hat{d}_i} - \log d_i)^2 - \frac{1}{n^2} \sum_{i,j} ( (\log {\hat{d}_i} - \log d_i) (\log {\hat{d}_j} - \log d_j) )
\end{equation}{}
        
\textbf{Equation \ref{eqn:eigenloss1},} subtracts the mean loss from the loss of each pixel in order to make up for mistakes due to scale ambiguity. We could have a different interpretation by reformulating it as \textbf{Equation \ref{eqn:eigenloss2},} which compares each pixel pair in the ground truth and the same pixel pair in the estimated depth map, in order to achieve the same distance between pixels in both the ground truth and the estimated depth maps. Similarly, the same loss function can be written as in \textbf{Equation \ref{eqn:eigenloss3},} which can be interpreted as the network being penalized for mistakes in different directions and being rewarded for mistakes in the same direction.
        
All three above formulations are equivalent, and actually a fourth reformulation where the function can be computed in linear time, is used to train the network:
\begin{equation}
    \label{eqn:eigenloss4}
    loss = \frac{1}{n} \sum_i (\log {\hat{d}_i} - \log d_i)^2 - \frac{\lambda}{n^2} ( \sum_i (\log {\hat{d}_i} - \log d_i) )^2
\end{equation}{}
Here, $\lambda$ is a hyperparameter that controls the scale invariance of the loss. $\lambda=1$ refers to fully scale-invariant, and $\lambda=0$ refers to the normal L2 loss.
         
Following their previous work in \cite{eigen_depth_2014} very closely, and building on top of it, Eigen et al. \cite{eigen_predicting_2015} devise a network architecture that can successfully estimate depth, surface normals, and semantic labels. Note that they do not optimize their system jointly for the said tasks, instead, they merely show that a single neural network architecture can solve all the tasks. Although they experiment with shared layers between depth and surface normals, it does not improve their results.
        
As an improvement to their previous work, another scale is added to their multi-scale network architecture, which further refines the output using convolutional layers. Additionally, they incorporate an extra term to their loss as shown in \textbf{Equation \ref{eqn:eigenloss5},} which tries to match the gradient of the depth in the estimated depth map and the ground depth map, that results in a better local structure in the output depth maps: 
\begin{equation}
    \label{eqn:eigenloss5}
    loss = \frac{1}{n} \sum_i y_{i}^2 - \frac{\lambda}{n^2} \left( \sum_i y_{i} \right)^2 
    + \frac{1}{n} \sum\left[\left(\nabla_{x} y_{i}\right)^{2}+\left(\nabla_{y} y_{i}\right)^{2}\right],
\end{equation}{}
where $y_i = \log {\hat{d}_i} - \log d_i$ and $\nabla_{x} y_{i}$ and $\nabla_{y} y_{i}$ are image gradients.

Here we see two important themes:
\begin{boitecoloriee}

    \textbf{Custom loss functions} While the handcrafted features are replaced with learned features by the neural networks, we see problem specific knowledge being embedded into the solutions as custom loss functions in many works. While the scale-invariant loss function \cite{eigen_depth_2014} tries to alleviate the negative effects of the scale ambiguity during training, the gradient matching term \cite{eigen_predicting_2015} helps the model to preserve local structure in the depth map.
    
    \textbf{Learning multi tasks} The reason many scientists tackled the SIDE problem by learning multiple tasks \cite{hoiem_closing_2008, liu_single_2010, karsch_depth_2014, ladicky_pulling_2014, eigen_predicting_2015, wang_towards_2015, li_depth_2015, yan_single_2017, ren_cross-domain_2018, kim_unified_2016, qi_geonet_2018, xu_pad-net_2018, zhang_joint_2018, zhang_pattern-affinitive_2019, chen_towards_2019, lu2020taskology} is twofold. First, estimating depth from a given single RGB image requires a high-level understanding of the scene such as detecting and recognizing objects and their relations in 3D. However, the loss from depth estimation alone may not be enough for the network model to discover those high-level relations. Therefore, additional tasks such as surface normal estimation, semantic segmentation, intrinsic image decomposition along with their corresponding losses are utilized to train the networks. Moreover, we know that some of the aforementioned tasks share a common understanding of the given scene; therefore, it is beneficial to learn them jointly to increase the robustness of the network.
\end{boitecoloriee}
        
Both \cite{eigen_depth_2014} and \cite{eigen_predicting_2015} were able to achieve very good results at that time. Nevertheless, there were still opportunities for improvement. First of all, most real-world tasks require a real-time vision pipeline, and Eigen et al.'s framework \cite{eigen_depth_2014} is slow. Another of its weaknesses is having too many learnable parameters that lead to an increase in memory requirements, and the need for too many training points to train the network. Lastly, the resolution of the output is very low. Again, from the perspective of the application, higher resolution outputs are more desirable.
        
Laina et al. \cite{laina_deeper_2016} aim to solve these problems in their work by applying recent developments in CNN technology. Essentially, they follow the work of Eigen et al. \cite{eigen_depth_2014} by training a CNN to regress per pixel depth value. Instead of having two networks, and fully connected layers, they train a fully convolutional residual network that is based on ResNet-50 \cite{he_deep_2016}, with up-convolutional blocks at the end. Their residual network has fewer parameters, runs in real-time, and up-convolutional blocks increase the resolution of the output. Even though the lack of fully connected layers looks like a problem in terms of global context, the receptive field of the deep residual network covers the whole input image, providing the necessary global context.
        
Another improvement of \cite{laina_deeper_2016} is on the loss function. After observing a heavy-tailed distribution of depth values in the datasets, it uses a reverse Huber loss called BerHu, which acts as an L1 loss below threshold $c$ and acts as an L2 loss above that threshold. The BerHu loss is given by
\begin{equation}
\label{eqn:lainaloss1}
    loss( d_i, \hat{d_i} )=\left\{\begin{array}{ll}
    |d_i - \hat{d_i}| & |d_i - \hat{d_i}| \leq c \\
    \frac{(d_i - \hat{d_i})^{2}+c^{2}}{2 c} & |d_i - \hat{d_i}|>c
    \end{array}\right.,
\end{equation}
where $c$ is calculated over all the pixels of a batch of input images as $c=\frac{1}{5} \max _{i}(|d_i-\hat{d_i}|)$. This loss helps to put more emphasis on small residuals while still having the advantage of L2 loss for high residuals.
        
Cao et al. \cite{cao_estimating_2017} formulate the SIDE problem as a classification problem. The main motivation for this is twofold. First of all, it is hard to regress to the exact depth value, even humans have a hard time estimating the exact depth. Instead, one can estimate the depth range with ease. Additionally, doing classification naturally produces a useful by-product that cannot be produced by doing regression without extra difficulty. The by-product is the confidence on the estimation that can be utilized to further enhance the estimation of the network by updating the network's estimation with the low confidence based on neighboring estimations with high confidence as post-processing.
        
Cao et al.'s framework \cite{cao_estimating_2017} consist of two parts. The first part is a ResNet based on a fully convolutional network that estimates the depth range for each pixel in the input RGB image. The depth ranges are acquired by uniformly discretizing the continuous depth values in the log space in the ground truth depth maps. This network is trained with a cross-entropy loss weighted by an information gain matrix so that depth ranges close to the ground truth depth range are also used to update the network's weight as follows: 
\begin{equation}
    \label{eqn:caoloss1}
    loss=-\frac{1}{N} \sum_{i=1}^{N} \sum_{D=1}^{B} H\left(D_{i}^{GT}, D\right) \log \left(P\left(D | z_{i}\right)\right)
\end{equation}
where $H(p, q)=\exp \left[-\alpha(p-q)^{2}\right]$, and $\alpha$ is a constant. $D$ refers to discrete depth labels while $D_i^{GT}$ refers to ground truth depth label for $i$\textsuperscript{th} image, $P$ refers to estimated probability for the depth label. Normally, the cross-entropy loss tries to increase the probability of the correct class. Here, this loss also tries to increase the probability of the closer classes in the depth domain. In the test time, the center of the estimated depth range is assigned to a pixel's depth value. 
        
The second part of Cao et al.'s framework is a fully connected conditional random fields (CRF) \cite{lafferty_conditional_2001} that consists of unary potentials for each pixel and pairwise potentials for each pixel pair in the image. While the unary potential pushes the system to output correct labels for each pixel, the pairwise potential smooths the depth estimation by looking at pixels' positions and their appearance in the RGB domain.

An important theme we see here is:
\begin{boitecoloriee}
    \textbf{Classification and ordinality} A number of works decided to formulate the depth estimation problem in a way that does not require the system to estimate the exact depth value \cite{zoran_learning_2015, cao_estimating_2017, fu_deep_2018, chen_single-image_2016, xian_monocular_2018, chen_learning_2019, mertan2020relative, mertan2020new, xian2020structure, lienen2020monocular}. While estimating the relative depth instead of the absolute depth is an alternative, estimating the depth range in a classification setting can be done with success. It is important to see that the classification formulation of the depth estimation is inherently different than the classic classification formulation since the classes in the depth estimation problem represent depth ranges that have ordinal relations with one another. This has been utilized to increase performance \cite{cao_estimating_2017, fu_deep_2018}.   
\end{boitecoloriee}
            
Fu et al. \cite{fu_deep_2018} determine two important drawbacks of the current approaches so far. Similar to \cite{cao_estimating_2017}, they advocate that the regression formulation of the depth estimation is hard to train with, thus results in poor solutions. Additionally, they claim that the networks that have been used consist of spatial pooling operations, which reduce the output resolution, followed by deconvolutions and skip connections to increase the resolution, which, as a whole, complicates the networks and adds computational cost. 
            
To improve upon their observations, they formulate the problem as an ordinal regression problem. Despite the name having regression in it, their system actually consists of a network with multiple heads where each head of the network solves a pixel-wise binary classification problem in which the aim is to decide whether a pixel is closer or further away from a particular depth threshold. More specifically, they pick the number of depth thresholds $t_i \in t_0, t_1, .., t_{K-1}$ defined by: 
\begin{equation}
    \label{eqn:SID}
    t_{i}=e^{\log (\alpha)+\frac{\log (\beta / \alpha) * i}{K}},
\end{equation}
where $ [\alpha, \beta] $ is the depth interval. 
The important thing to notice here is that it increases the distance between consecutive thresholds, which affects the loss in a way which working in the log space does in classic regression formulation.
            
The network has a head $h_{t_i}$ for each of the $t_i$s, and each head classifies all the pixels. The network is trained with the loss:
\begin{equation}
    \label{eqn:DORNloss}
    loss =-\frac{1}{N} \sum_{i=0}^{N-1} \sum_{k=0}^{K-1} 
    [d_{i} > t_i] \log(1 - \hat{y}_{i}^k) + [d_{i} < t_i] \log(\hat{y}_{i}^k),
\end{equation}
where $N$ is the number of pixels, $[.]$ is the indicator function, $d_i$ is the ground truth depth of the i\textsuperscript{th} pixel, and the $\hat{y}_i^k$ is the prediction of the k\textsuperscript{th} head for the i\textsuperscript{th} pixel.
            
Inference stage also needs special care as we have multiple ordinal predictions for each pixel. The estimated depth of a pixel is calculated using the classification probabilities obtained by each head of the network for that pixel with the formula given by:
\begin{equation}
    \label{eqn:DORNinference}
    \begin{aligned}
    &\hat{l}_{(w, h)}=\sum_{k=0}^{K-1} [\hat{y}_{i}^{k}>=0.5]\\
    &\hat{d}_{i}=\frac{t_{\hat{l}_{i}}+t_{\hat{l}_{i}+1}}{2}.
    \end{aligned}
\end{equation}
            
Following the trends of the time of the paper \cite{fu_deep_2018}, they replace spatial pooling operations with dilated convolutions, which allow them to preserve the resolution of the image and get rid of costly up-sampling parts of previous architectures. Their architecture consists of a generic feature extractor part followed by a scene understanding modular which, in parallel, applies dilated convolutions with different dilation rates, 1x1 convolutions to detect cross channel features, and a full image encoder to capture global information. All the feature maps created by the scene understanding module are concatenated and fed to the heads of the network for final pixel-wise prediction.

Bhat et al. \cite{bhat2020adabins} improve upon previous solutions by more explicitly utilizing global information. They make two key observations. First, they observe that depth values in a dataset, even when it is an indoor dataset such as NYU, varies greatly from image to image. While the depth distribution of a close up image of furniture is mostly focused on smaller values, the depth distribution of an image of a hallway may be focused on bigger values. Additionally, they observe that previous methods gather global information through successive convolutional layers, which decreases the spatial resolution of the image before achieving a big receptive field over the input image. They conjecture that this results in poor exploitation of global information. 

Based on their observations, they propose to learn adaptive bins per image. Similar to the ordinal regression idea of Fu et al. \cite{fu_deep_2018}, they divide the depth range into N bins and estimate the probability of bins per pixel. For an input image, their model outputs a vector that can be used to determine the depth range of each bin and a per-pixel probability map that shows the likelihood of each bin. The final depth of a pixel is calculated as a linear combination of bin centers weighted by the probabilities, as opposed to choosing the center of the most likely bin as in \cite{fu_deep_2018}. This results in smoother depth maps, which are preferred for the downstream tasks. Additionally, choosing the bins adaptively based on the input image allows the model to easily focus on certain depth ranges and increase the overall depth estimation performance.

Adopting the recent developments in deep learning, they use a variant of Vision Transformer in their architecture. Their model is trained with a scaled version of scale-invariant loss of Eigen et al. \cite{eigen_depth_2014}. Additionally, they use the chamfer distance between estimated bin centers and ground truth depths as a regularizer to provide an additional supervision signal for the estimation of the bin center. 
        
        \subsubsection{Multitasking}
        
        Here, we list the works that solve the SIDE problem jointly with similar task or tasks. In our case, similar means the closeness of tasks that are being solved jointly in the solution domain. As we discussed earlier, synergizing similar tasks helps to improve the results and increase the robustness of the acquired solutions. \textbf{Table~\ref{tab:multitask}} summarizes the works that we discuss in the section. Note that this list does not contain all the works utilizing multitasking as some of the works are discussed under different sections to emphasize different aspects of the said works.

\begin{table*}[]
\centering
\caption{\label{tab:multitask}Summary of the works that combine SIDE with similar tasks.}
\begin{adjustbox}{width=1\textwidth}
\begin{tabular}{lllll}
 & \begin{tabular}{c}auxiliary task\end{tabular} & \begin{tabular}{c}method\end{tabular} & \begin{tabular}{c}framework\end{tabular} & \begin{tabular}{c}key point\end{tabular} \\ \hline\hline
 
\centered{Liu~et~al.~\cite{liu_single_2010}} & \centered{semantic segmentation} & \centered{MRF} & \centered{two-staged} & \centered{Semantic classes can be exploited for depth estimation.} \\\hline

\centered{Ladicky~et~al.~\cite{ladicky_pulling_2014}}& \centered{semantic segmantation} & \centered{multi-class boosted \\classifier} & \centered{separate classifier for\\ each semantic class} & \centered{It is better to estimate whether an object is at an \\arbitrarily fixed canonical depth since the look of an \\object varies greatly depending on the depth.} \\\hline

\centered{Qi~et~al.~\cite{qi_geonet_2018}}& \centered{surface normal estimation} & \centered{ResNet based CNN} & \centered{two-staged} & \centered{Both depth and surface normal estimations can be \\refined by using each other as they are geometrically \\related under the assumption of neighboring pixels lie \\in the same local tangent plane in 3D.} \\\hline

\centered{Xu~et~al.~\cite{xu_pad-net_2018}} & \centered{semantic segmentation +\\ surface normal estimation +\\ contour estimation} & \centered{ResNet based CNN} & \centered{two-staged} & \centered{Pseudo multi-modal input can be created by solving \\auxiliary intermediate tasks.} \\\hline

\centered{Zhang~et~al.~\cite{zhang_joint_2018}} & \centered{semantic segmentation} & \centered{ResNet based CNN} & \centered{task-recursive} & \centered{High-level cross-task knowledge can be utilized by \\switching between tasks throughout the framework.} \\\hline

\centered{Zhang~et~al.~\cite{zhang_pattern-affinitive_2019}} & \centered{semantic segmentation +\\ surface normal estimation} & \centered{ResNet based CNN} & \centered{two-staged} & \centered{Similarities in feature space can explicitly be calculated \\and incorporated for further refinement of each task.} \\\hline

\centered{Chen~et~al.~\cite{chen_towards_2019}} & \centered{semantic segmentation} & \centered{DispNet~\cite{mayer2016large} variant\\CNN} & \centered{task-conditional} & \centered{The decoder can be conditioned with a task identity \\to allow model to share knowledge to a greater extent.} \\ \hline

\centered{Lu~et~al.~\cite{lu2020taskology}} & \centered{semantic segmentation +\\ ego-motion / \\surface normal estimation} & \centered{UNet with ResNet18\\ base} & \centered{distributed} & \centered{One can train models for each task separately and\\ combine them just by using a consistency loss derived \\from physical and logical constraints.} \\ \hline

\end{tabular}
\end{adjustbox}
\end{table*}
        
Liu et al. \cite{liu_single_2010} combine SIDE with the semantic segmentation task. As explained earlier, the importance of the semantic information for depth estimation task is their main insight. They design a two-stage system where semantic labels are predicted and used to apply additional constraints for the depth estimation task. Furthermore, the SIDE problem becomes easier to solve as simpler features may explain the depth of a point conditioned on the semantic class information. 

MRFs are utilized to estimate the per-pixel semantic classes by minimizing a unary label potential and a binary smoothing potential that is guided by similarity in the RGB domain. On top of semantic segmentation, they estimate the horizon, and semantically decompose the scene, which is similar to \cite{hoiem_automatic_2005}. This decomposition is used to compute additional hand-crafted constraints on the depth of the pixels. To be able to fully exploit available semantic information, a separate MRF model for each semantic class is trained to estimate the depth in the second stage. 
        
Ladicky et al. \cite{ladicky_pulling_2014} observe that older approaches do not incorporate perspective information. The look of an object in the images varies greatly depending on its depth and it is a problem for data-driven approaches as they need samples for each object at each possible depth to be able to learn accurate representations. They claimed that this is an important problem for both semantic segmentation and depth estimation tasks.

To overcome this problem, they use the fact that the depth of a pixel in an image will change reversely proportionately to any scaling of the image. Explicitly, the expression: 
\begin{equation}
    \label{eqn:scaling}
    H_d(I, i) = H_{d / \alpha}(\alpha * I, \alpha i)
\end{equation}
 holds for every pixel $i$, and scaling factor $\alpha$, where $d$ is an arbitrary depth, $H_d(I,i)$ is the probability of pixel $i$ to be at depth $d$ in image $I$, and $\alpha * I$ is the geometrically scaled image.
 
Using this information, they train $L$ number of classifiers $H_{d, l}(I,i)$ that output the probability of a pixel $i$ in image $I$ being at an arbitrarily chosen depth $d$, and having the semantic class label $l$ where $L$ is the total number of semantic classes. The input image $I$ can be scaled with different scaling factors $\alpha$ to determine the correct scaling that projects the pixel to the chosen depth $d$. The actual depth of the pixel can then be calculated using the obtained scaling factor. This reformulation of the problem helps the system by reducing the joint problem of estimating the depth and semantic class into the simpler problem of detecting whether a pixel is of a particular depth and of a particular class. They claim that the learned features would be simpler since they only have to work for a specific depth as opposed to features that have to work across a depth range.
        
Wang et al. \cite{wang_towards_2015} try to unify SIDE and semantic segmentation in a framework by jointly solving both tasks instead of solving them sequentially. The latter is prone to error due to propagation from one task to another. Their framework consists of a complicated hand-crafted pipeline, which includes a CNN for pixel-wise depth and semantic label prediction, another CNN for a super-pixel based depth and semantic label prediction, and a hierarchical CRF that works on both pixel level and super pixel-level estimations, and refines them by applying numerous handcrafted constraints similar to the constraints we have seen so far in other works. While the CNNs estimate depth and semantic labels on a global scale, hierarchical CRF refines the results locally. They were able to show that joint training of depth and semantic segmentation increases the performance of both tasks individually. 

Qi et al. \cite{qi_geonet_2018} tackle the SIDE problem by combining it with surface normal estimation. Their network is called GeoNet, a geometric network because they are utilizing geometric relations between depth and the surface normals to refine their estimations. Their main observation is the inability of neural networks to learn these geometric relations from data directly, which they have shown by the poor performance of a CNN on the task of estimating surface normals from given depth maps.

The main assumption to allow applying geometric constraints is that the neighboring pixels lie in the same local tangent plane in 3D. This assumption allows us to fit a local tangent plane in a neighborhood of pixels using their depths or estimating the depth of pixels in the neighborhood using surface normals assuming that they are on the same local tangent plane. This geometric relation between depth and surface normals is utilized to boost the performance of both tasks.

Qi et al.'s framework \cite{qi_geonet_2018} consists of a generic network that makes the initial estimation for depth and the surface normals given the RGB image. Afterward, these initial estimations are refined by "depth-to-normal" and "normal-to-depth" networks while both of the networks take the initial estimations as input.

The "Depth-to-normal" network takes the estimated initial depth values and calculates the 3D position of each pixel. The neighborhood for any pixel $N_i$ is calculated based on the closeness in coordinates in 2D, and closeness in the  depth value, where the closeness is controlled by hyperparameters $\beta$ and $\gamma$ as given by: 
\begin{equation}
    \label{eqn:geonet_n_d}
    \begin{array}{c}
    N_{i}=\left\{\left(x_{j}, y_{j}, z_{j}\right) |~~ |u_{i}-u_{j} |<\beta\right., \\
    \left.\left|v_{i}-v_{j}\right|<\beta,\left|z_{i}-z_{j}\right|<\gamma z_{i}\right\},
    \end{array}
\end{equation}
where $(u,v)$ represents the coordinate of a pixel in the image domain. 
Using the aforementioned assumption, a local tangent plane is fitted to each neighborhood, and then surface normals are calculated for each pixel. Note that these steps can be done deterministically, without the need for learning any parameters. Subsequently, a residual CNN module is applied to fuse the geometrically calculated normals and the initial normal estimation and produces the refined end result for surface normals.

The "Normal-to-depth" network works in a similar fashion. It determines the local neighborhood by the distance in the 2D image coordinates and the surface normals in the initial estimation. For each pixel $j$ in the neighborhood of pixel $i$, a depth estimation for pixel $i$ is calculated based on the assumption that they are lying on the same local tangent plane. Subsequently, these assumptions are aggregated using a kernel regression with a linear kernel. Overall, the initial depth estimation is refined using the initial normal estimation and geometric constraints without introducing any learnable parameters to the framework.

Xu et al. \cite{xu_pad-net_2018} tackle the SIDE problem jointly with the semantic segmentation problem, by predicting and combining intermediate outputs consisting of complementary tasks. Their main idea is to utilize a pseudo multi-modal input for the estimation of each single task since we expect a network to perform better with multi-modal inputs compared to inputting just an RGB image.

To this end, they devise a two staged framework. In the first stage, a generic encoder $E$ generates feature maps that are processed by separate decoders $D_d, D_{sf}, D_c, D_{ss}$ to output pixel-wise depth $d_{intermediate}$, surface normals $sf_{intermediate}$, contour labels $c_{intermediate}$, and semantic labels $ss_{intermediate}$.
        
In the second stage, a multi-modal distillation module is devised to combine the intermediate outputs to predict depth estimation and semantic segmentation separately. First, intermediate outputs are turned into feature maps $f_{t}$ where $t \in \{d, sf, c, ss\}$ by expanding the channel dimensions via applying convolutional layers. For the depth estimation task, an attention map $a_{d}$ is produced using feature maps of intermediate depth output $f_d$, and feature maps of intermediate complementary tasks are gated with this attention map. Gated feature maps and feature maps of the intermediate depth output then are concatenated and passed to the final prediction module $H_{d}$ that consists of convolutional layers to estimate the final depth map $d_f$ as follows:
\begin{equation}
    \label{eqn:padnet}
d_f = H_d\left( cat( f_{d}, a_{d} \otimes f_{sf}, a_{d} \otimes f_{c}, a_{d} \otimes f_{ss}   ) \right),
\end{equation}
where $\otimes$ represents pixel-wise multiplication. 
Similarly, feature maps of the intermediate semantic output used for attention mechanism and combined feature maps are passed to another module to predict final semantic labels.

A simple Euclidean loss for estimation of depth and surface normals, a cross-entropy loss for contour detection, and softmax loss for semantic segmentation are used for all intermediate and final outputs. The whole architecture is trained in an end-to-end fashion with a linear combination of these losses.

Zhang et al. \cite{zhang_joint_2018} also tackle the SIDE problem with semantic segmentation jointly. However, they bring a different perspective to the multi-task learning problem. Instead of having an architecture with shared layers up to a point and bifurcating from there on, they have devised a framework inspired by the human learning system where the framework alternates between the two complementary tasks back and forth. This allows them to utilize high-level cross-task knowledge, compared to sharing layers where just common information is shared between tasks.
        
Zhang et al.'s architecture is an autoencoder with skip connections from the encoder part where the decoder part of the network outputs multi-scale depth and semantic segmentation estimations. The decoder consists of four sequential main blocks $D^i \in D^1, D^2, D^3, D^4$, each of which has heads $H^i_d, H^i_s$ to output depth and semantic segmentation predictions $d^i, s^i$ in different scales. Each block of the decoder consists of two inner blocks $D^i_d, D^i_s$ where the feature maps generated by the first inner block are used to output depth and the feature maps generated by the second inner block are used to estimate semantic labels. Furthermore, feature maps of previous depth and semantic segmentation outputs are gated with modules called $TAM$ which applies the attention mechanism and used as an input for the higher scales. \textbf{Equation \ref{eqn:TRLdecoder}} summarizes the process. Note that skip connections from the encoder part of the architecture are omitted for brevity.
\begin{equation}
    \label{eqn:TRLdecoder}
    \begin{aligned}
        feature_d^i & = D^i_d\left( TAM(feature_s^{i-1},feature_d^{i-1}) \right)\\
        d^i & = H^i_d( feature_d^i )\\
        feature_s^i & = D^i_s\left( TAM(feature_s^{i-1},feature_d^i) \right)\\
        s^i & = H^i_s( feature_s^i )
    \end{aligned}
\end{equation}
Essentially, their framework refines the depth and semantic segmentation outputs, going from smaller scales to bigger scales, alternating between depth estimation and semantic segmentation tasks utilizing the cross-task knowledge similar to \cite{xu_pad-net_2018}.

Zhang et al. \cite{zhang_pattern-affinitive_2019} combine SIDE, surface normal estimation, and semantic segmentation tasks in their framework. They point out an important issue of multi-task approaches: ambiguity in the feature learning. Essentially, they claim that trying to learn features for multiple tasks may create ambiguity. Instead of letting a network learn the features that may be useful for all the tasks at hand, they propose to devise a method where they can explicitly look for similar features between tasks.

First of all, they investigate the existence of such patterns between tasks. To this end, they label pixel pairs as similar if their depth difference is smaller than a threshold or dissimilar if vice versa. The same process is applied to surface normal ground truths and semantic segmentation maps in which the similarity is decided based on the label equality. They have shown that 50\% - 60\% percent of pixel pairs that are labeled as similar in one task are similar in other tasks and the same holds for dissimilar pairs too.

To exploit this observation, they devise a pattern-affinitive propagation method. Firstly, the RGB image is processed by a shared encoder, and task-specific decoders output initial estimates for each task. Using the last feature maps $f_d, d_{sf}, f_{ss}$ generated by task-specific decoders, affinity matrices $M_d, M_{sf}, M_{ss}$ are calculated as 
\begin{equation}
    \label{eqn:PAPaffinity}
    M_{t, ij} =  e^{- s(f_{t, i}, f_{t, j}) },
\end{equation}
where $s()$ is the similarity function such as an L1 distance, inner product, and so on. Additionally, each row of the affinity matrix is normalized. Basically, it expresses the similarity probability in the feature space for each pixel pair $ij$. Note that the calculation of affinity matrices is done deterministically and does not introduce any learnable parameters. After the affinity matrices are calculated, they are combined by a linear weighting using a separate learnable weight set for each task,  which is used to propagate affinities in the feature space. The final predictions are done using the affinity propagated features.
        
Chen et al. \cite{chen_towards_2019} tackle unsupervised SIDE and supervised semantic segmentation problems together. Unsupervised depth estimation is performed similarly to the stereo approach of Godard et al. \cite{godard_unsupervised_2017}. Similar to previous works, they propose architectural improvements to have a better multi-task learner as well as additional losses that utilize cross-task knowledge to boost the performance of the tasks.

Their proposed architecture consists of an encoder $E$, which produces a scene representation $z$ of the given RGB image $I$, and a decoder network $D$ with skip connections from the encoder, which takes the scene representation $z$ and the task identity $t$ to generate cross-modal prediction $\hat{y}$. Task identity $t$ is just a layer with all 1s for the depth estimation task and all 0s for the semantic segmentation task. Essentially, the same decoder learns to generate predictions conditioned on the task identity, allowing them to share weights and knowledge to a greater extent. Lastly, either softmax or pixel-wise average pooling is applied to cross-modal prediction to get semantic class probabilities or the depth map, respectively. \textbf{Equation \ref{eqn:towards}} summarizes the framework:
\begin{equation}
    \label{eqn:towards}
    \begin{aligned}
    z &= E(I),\\
    \hat{y} &= D( cat( z, t ) ),\\
    \hat{y}_{final} &= f( \hat{y} ),
    \end{aligned}
\end{equation}
where $f$ is the softmax function if the task is semantic segmentation, and pixel-wise average pooling if the task is depth estimation.
                
Furthermore, two self-supervised losses are defined to constrain the depth estimation using semantic knowledge. Similar to the reconstruction of one stereo RGB image from another using the estimated depth information, estimated semantic segmentation maps are warped to reconstruct each other, and the L1 loss is applied to penalize inconsistencies between the reconstructed semantic segmentation maps. Additionally, a regularization loss that is guided by semantic class predictions is added since, intuitively, similar depth values within the same object are desired.

Lu et al. \cite{lu2020taskology} describe a general framework, named Taskology, for scalable, modular multi-task learning. In Taskology, combined tasks are trained separately in a traditional single-task fashion, using their own architectures, losses, and datasets. What they share is an additional "consistency loss", derived from physical and logical constraints that connect the tasks. It measures how consistent estimations of each task with each other. Since the consistency loss does not require ground truth information, it can be calculated for inputs coming from an additional dataset that doesn't have to have labels for any of the tasks. 

In particular, Taskology is applied to predict depth, ego-motion, and semantic segmentation together, as well as depth and surface normals. In the first case, predicting the depth, ego-motion, and semantic segmentation allows us to reconstruct the next image in a stereo pair or a neighboring frame in a video. Hence, photometric loss between the original image and the reconstructed image is used as a consistency loss. The same logic can be applied to the segmentation maps as well. Predicted segmentation maps for the neighboring images should be the same when they are warped onto each other. In the second case, surface normals can be calculated from depth predictions. The difference between predicted surface normals and the surface normals calculated from predicted depth maps is used as a consistency loss. Moreover, both the depth and the surface normal predictors are trained on synthetic data while the consistency loss is applied to the predictions of real images. Their experiments show that applying consistency loss on real images helps models to bridge the domain difference. 
    
    \subsection{Main consideration: eliminating the need for labeled data}
    In this research category, researchers try to eliminate the need for labeled data for the SIDE problem as labeled data is hard to acquire. Additionally, the need for labeled data prevents the life long learning opportunity, makes it hard to fine-tune networks for unique situations. Two approaches have been developed by the researchers, namely usage of synthetic data and unsupervised learning without any labeled data. The works we discuss below are summarized in \textbf{Table~\ref{tab:unsupervised}}.
    
    \begin{table*}[]
    \centering
    \caption{\label{tab:unsupervised}Summary of the works that do not use annotated real world ground truth information. }
    \begin{adjustbox}{width=1\textwidth}
    \begin{tabular}{llll}
     & \begin{tabular}{c}supervision source\end{tabular} & \begin{tabular}{c}loss function\end{tabular} & \begin{tabular}{c}key point\end{tabular}  \\ \hline\hline
     
     \centered{Ren~et~al.~\cite{ren_cross-domain_2018}} & \centered{synthetic images} & \centered{scale-invariant loss} & \centered{Synthetic images with multiple ground truth annotations for different \\tasks can be used for training and domain adaptation can be applied \\to bridge the gap between real and synthetic images.} \\ \hline
     
     \centered{Garg~et~al.~\cite{garg_unsupervised_2016}} & \centered{stereo images} & \centered{photometric loss + L2 regularization \\on disparities} & \centered{Estimated depth information can be used to reconstruct different \\views of the same scene. The reconstruction loss creates sufficient \\supervision to learn to estimate the depth information.  } \\ \hline
     
     \centered{Godard~et~al.~\cite{godard_unsupervised_2017}} & \centered{stereo images} & \centered{photometric loss + edge-aware L1 \\regularization on disparities + left\\right consistency loss on disparities} & \centered{Having two images of the same scene with known camera baseline \\allows us to define a consistency loss, which improves the performance.} \\ \hline
     
     \centered{Zhou~et~al.~\cite{zhou_unsupervised_2017}} & \centered{monocular video} & \centered{masked photometric loss + L1 \\regularization on second order \\gradients of the depth} & \centered{The reconstruction loss is also sufficient for learning to estimate the \\camera motions between subsequent frames of a monocular video. \\Estimated camera motions allow monocular videos to be utilized \\instead of stereo images. } \\ \hline
     
     \centered{Wang~et~al.~\cite{wang_learning_2018}} & \centered{monocular video} & \centered{photometric loss + edge-aware L1 \\regularization on second order \\gradients of the depth} & \centered{Camera motion estimation can be replaced with direct visual odometry. } \\ \hline
     
     \centered{Ranjan~et~al.~\cite{ranjan_competitive_2018}} & \centered{monocular video} & \centered{photometric loss + edge-aware L1 \\regularization on first order \\gradients of the depth} & \centered{Tasks that are geometrically related such as SIDE, camera motion \\estimation, optical flow, and motion segmentation, can be solved jointly \\in an unsupervised fashion with a three player game. } \\ \hline

    \end{tabular}
    \end{adjustbox}
    \end{table*}

        \subsubsection{Usage of synthetic data}
        
        One of the ways to overcome the need for labeled data is the usage of synthetic data. Even though the training is done in a supervised way with labeled data, synthetic data eliminates the costly real data collection since synthetic data can be created in an automatic way in great amounts and in great diversity.
        
Ren and Lee \cite{ren_cross-domain_2018} try to solve the SIDE problem by learning more than one complementary task. Here they are inspired by humans as humans usually learn things jointly, utilizing more than one source of information. In particular, they try to estimate depth, surface normals, and instance contours at the same time. They believe learning these tasks jointly would increase the performance of the system as the tasks share some common understanding for a given scene. However, it is very hard to find a dataset that has labels for all of the tasks at hand. Therefore, they created their own data synthetically which allowed them to create a big enough dataset with all the needed ground truths. They trained an estimator network, which is a CNN with three heads to estimate depth, surface normals, and instance contours jointly.
        
One problem that the authors \cite{ren_cross-domain_2018} face is the domain difference between the real and synthetic images. Since they do not assume labels for real data, they force the network to have similar low-level feature maps for both real and synthetic images by applying adversarial training. Essentially, a discriminator network is trained for discriminating between synthetic and real-world images by looking at low-level feature maps produced by the estimator network. Afterward, an adversarial loss is produced from the discriminator network and it is used to train the estimator network so that the gap between synthetic and real images in low-level feature maps diminishes.

        \subsubsection{Unsupervised learning}
        
        Another way of truly overcoming the need for labeled data is unsupervised learning. Unsupervised learning for the SIDE problem uses stereo images or video recordings with small changes in camera positions between frames as two consequent frames can be considered as stereo images. All the unsupervised methods roughly work as follows: the system takes a pair of RGB images (let us call them $ I_{left} $ and $ I_{right} $) which show the same scene from two slightly different perspectives. Then the depth is estimated for one of the images, for instance, let us assume that we estimate the depth of $ I_{left} $, as $ D_{left}$. Using the estimated depth and the camera motion between frames, other image in the pair can be deterministically warped to produce the first image using visual geometry as follows:
\begin{equation}
    I_{right}(D_{left}) \sim I_{left}.
\end{equation}{} 
Afterward, the network can be trained with the reconstruction error. 
\begin{equation}
    loss = f(I_{right}(D_{left}) - I_{left}). 
\end{equation}{}
Fundamentally, the system trains in a self-supervised way where it tries to produce the input image itself as an output. Meanwhile, depth maps are produced by the system as a middle step and in order to minimize the reconstruction loss, the system learns how to predict more accurate depth maps. 
        
This method makes some assumptions. First of all, as it assumes static scenes; objects' positions in 3D should be the same in $ I_{left} $ and $ I_{right} $. Lack of occlusion/disocclusion is also assumed. Lastly, it assumes Lambertian surfaces, i.e. reflectance of the surfaces does not change with the point of view. Even though these assumptions may not hold all the time, researchers were able to successfully train networks in an unsupervised way.
        
To the best of our knowledge, the first work that tackles the SIDE problem in an unsupervised way is the work of Garg et al. \cite{garg_unsupervised_2016}. They collect their own dataset using a calibrated stereo gig, which allows them to know the camera motion between two frames. Since they do not specifically aim for the best possible results, as stated in their work, they apply our basic definition of unsupervised learning of the SIDE problem. Just like an autoencoder where the image is first encoded into a smaller space and then decoded into itself with a minimal loss, their end-to-end trained fully convolutional network encodes the input image into a depth map and then inversely warps the other of the pair using the depth map with a reconstruction minimization loss. Note that the only part that is learned from data is the prediction of the depth map. The warping can be calculated directly and in a differentiable manner using the depth map and the known camera motion.
                 
Godard et al. \cite{godard_unsupervised_2017} follow the same basic principles and improve them. They train their network with rectified stereo pairs with known camera baselines to estimate disparity maps. The main difference from the previous work of Garg et al. \cite{garg_unsupervised_2016} is that the network produces left and right disparity maps ($ Disp_{left} $, $ Disp_{right} $) by looking at only one of the input images. Having left and right disparity maps allow us to reconstruct both of the images from each other,
\begin{equation}
    I_{left}(Disp_{right}) \sim I_{right}, ~~~~~
    I_{right}(Disp_{left}) \sim I_{left},
\end{equation}{}
and both of the reconstruction losses are used to train the network:
\begin{equation}
    loss = f( (I_{left}(Disp_{right}) - I_{right}) ) + 
    f( (I_{right}(Disp_{left}) - I_{left}) ) 
\end{equation}{}
where $f$ is the loss function consists of a weighted combination of SSIM and L1.
        
Additionally, they force their network to produce consistent disparity maps by penalizing the network for differences in disparity maps when they are projected onto each other:
\begin{equation}
    loss = \frac{1}{N} \sum_{i,j} | Disp_{left}^{(i,j)} - Disp_{right}^{( j + Disp_{left}^{(i,j)} )} |,
\end{equation}{}
where $ (i,j) $ is the coordinate of a pixel. Note that index into the disparity map varies in one dimension since the images are rectified.
        
Another work that does unsupervised learning of the SIDE problem is the work of Zhou et al. \cite{zhou_unsupervised_2017}. They train their network using a dataset consisting of video recordings. Videos in the dataset are acquired with a single camera, and subsequent frames are used to train the network as they have small camera position changes between them, hence they act like a stereo pair. Contrary to the previous work of Garg et al. \cite{garg_unsupervised_2016} and Godard et al. \cite{godard_unsupervised_2017}, they do not use known camera positions. Instead, they train an additional network to estimate the camera movement between frames. Therefore, they are able to use any video recording as an input to the training and eliminate the need for calibrated stereo image pairs. Simply, they estimate the depth $D_t$ of the target image $I_t$ with a network $DepthCNN$,
\begin{equation}
    D_{t} = DepthCNN( I_{t} ),
\end{equation}{}
estimate the camera motions $M_{t->t-1}$, $M_{t->t+1}$ between neighboring frames $I_{t-1}$, $I_{t+1}$ and the target frame $I_t$ with a network $PoseCNN$,
\begin{equation}
    M_{t->t-1}, M_{t->t+1} = PoseCNN( I_{t}, I_{t-1}, I_{t+1} ),
\end{equation}{}
and reconstruct the target image by warping the neighboring frames $I_{t-1}$ and $I_{t+1}$ using estimated camera motions $M_{t->t-1}$, $M_{t->t+1}$ and estimated depth $D_{t}$
\begin{equation}
    W( I_{t-1}, M_{t->t-1}, D_{t} ) = \sim I_{t},
\end{equation}{}
\begin{equation}
    W( I_{t+1}, M_{t->t+1}, D_{t} ) = \sim I_{t},
\end{equation}{}
where $W$ refers to warping function.
        
Another improvement to previously mentioned works is that they model the model limitation. As we discussed earlier, there are some assumptions that need to hold for this method to work. The authors train an additional network to predict how much their model can explain each pixel. In order to prevent bad pixels (pixels for which the assumptions do not hold such as a pixel on a moving car) having a negative effect on the training process, they weight the loss from each pixel using the predicted belief for that pixel:
\begin{equation}
    loss = ECNN( I_t ) \otimes f\left( I_t, W( I_{t-1}, M_{t->t-1}, D_{t} ), W( I_{t+1}, M_{t->t+1}, D_{t} )\right)
\end{equation}{}
where $ECNN$ is the network that produces a mask showing how much each pixel is explainable, $f$ is the photometric loss function, and $\otimes$ represents pixel-wise multiplication.
        
Wang et al. \cite{wang_learning_2018} point out two important weaknesses of the previous approach of Zhou et al. \cite{zhou_unsupervised_2017} and improve upon them. The first one is the scale ambiguity of the scene. As we discussed earlier, a change in the scale in the depth map results in the same 2D image; therefore, the photometric reconstruction losses of the two depth maps that only differ in scale will be the same. Since we train our networks with the guidance of the photometric reconstruction loss function, different scales of a scene are equally likely from the perspective of the network. The only difference here is the regularization term in the loss function which is used for the smoothness of the predicted depth map. Here the following problem occurs: as the scale of the scene decreases, the regularization loss also decreases. Therefore, the network learns to predict smaller and smaller scaled depth maps and eventually the training diverges. The authors successfully solve the problem by normalizing the output depth map before calculating the loss.
    
The second problem they detected is the separate estimation of depth and camera pose. They considered this as a problem since they believe these are related tasks. Additionally, with recent developments \cite{engel_lsd-slam_2014}, camera pose estimation between frames is treated as an optimization process by using frames and the depth, without any learning required, and in a differentiable manner. They injected this module into the framework and removed the pose estimation CNN. As a result, the depth estimating network is now updated with the gradients of the pose estimation module and the number of parameters to be learned is decreased.
        
Ranjan et al. \cite{ranjan_competitive_2018}, combine the idea of jointly solving related tasks and addressing the model limitation. Similar to Wang et al.'s work \cite{wang_learning_2018}, they solve related tasks together, namely SIDE, camera motion estimation, optical flow, and motion segmentation (segmentation of the scene into static and moving parts). As they state, all of these tasks are related, and solving them together benefits the system. Moreover, similar to the work of Zhou et al. \cite{zhou_unsupervised_2017}, they model the limitations of their model by segmenting the input image into moving and static parts and employing different networks to estimate the depth of different parts of the image. What makes their work interesting is that all of these tasks are learned through unsupervised learning, in a framework that they called "Competitive Collaboration".
        
In the Competitive Collaboration framework \cite{ranjan_competitive_2018}, there are three players and a resource. Two of the players compete for the resources, hence the name competitive. The last player, called the moderator, distributes the resources to the players, and to increase the overall performance of the competitors, it is trained by competitors, hence the name collaboration. In our case, the players are networks and the resource is the training data. The first network, $R$, estimates the optical flow for the static parts of the scene using depth and camera motion while the other competitor network, $F$, estimates the optical flow for moving parts of the image. The moderator network distributes the resource by segmenting the input image into static and moving parts which will be used to train corresponding networks. Competitors and the moderator take their turn in a training cycle that is similar to the expectation maximization method \cite{greff_neural_2017}. For details of the training procedure, application of it to a different problem, and theoretical analysis, refer to the paper, as they are not directly related to the SIDE problem.
    
    \subsection{Main consideration: working in the wild}\label{sec:itw}
    So far, the works that we summarized work on indoor and outdoor datasets separately. However, eventually, we would like to train systems that can work on images coming from unstructured environments, i.e. systems that work in the wild. \textbf{Table~\ref{tab:itw}} summarizes the works in this category.

    \begin{table}[]
    \caption{\label{tab:itw}Summary of the works that aim to work in the wild.}
    \begin{adjustbox}{width=1\columnwidth}
    \begin{tabular}{lccc}
    \hline
                    & loss             & sampling                          & network              \\ \hline \hline
    Zoran~et~al.~\cite{zoran_learning_2015} & softmax-loss     & -                                 & custom               \\
    Chen~et~al.~\cite{chen_single-image_2016} & pairwise ranking & -                                 & Hourglass            \\
    Xian~et~al.~\cite{xian_monocular_2018} & pairwise ranking & random~+~hard pairs                 & EncDecResNet               \\
    Chen~et~al.~\cite{chen_learning_2019} & pairwise ranking & random~+~hard pairs                 & EncDecResNet         \\
    Xian~et~al.~\cite{xian2020structure}  & pairwise ranking & random~+~instance edges~+~image edges & EncDecResNet         \\
    Mertan~et~al.~\cite{mertan2020relative} & listwise ranking & random                            & EncDecResNet         \\
    Lienen~et~al.~\cite{lienen2020monocular} & listwise ranking & random                            & EfficientNet variant \\ 
    Mertan~et~al.~\cite{mertan2020new} & distributional ranking & random & EncDecResNet \\\hline
    \end{tabular}
    \end{adjustbox}
    \end{table}

    There are two important issues to consider for depth in the wild problem. The first issue a system faces that aims to work in the wild is that the depth ranges may vary quite a lot which is a problem for the learning process. While the range of absolute depth values for an image of a table may be between 10 cm and a couple of meters, the image of a touristic place such as "Eiffel Tower" contains pixels that are a hundred meters away, if they have a depth value at all (sky regions can be considered infinitely far away). This kind of diversity makes the problem of estimating an absolute depth quite hard.

    Moreover, collecting datasets with absolute depth annotations for a diverse set of scenes that will allow a system to generalize unseen natural scenes is also very problematic. It requires a lot of effort in terms of time and money.
    
    It is clear that a new approach with new datasets was needed and researchers have quickly come up with a new formulation to the SIDE problem, which we refer to as relative-SIDE, which allows the system to generalize better in the wild. To this end, new datasets and new methods to learn within this problem setting were introduced. We summarize the datasets first and discuss the works that are done so far. 
    
    Before we start discussing the datasets, we want to summarize the work of Zoran et al. \cite{zoran_learning_2015}, which we see as the spiritual predecessor of the works we are going to discuss later on.
    
    Zoran et al. \cite{zoran_learning_2015} tackle the SIDE problem in a supervised manner. They formulate the problem as a three-way classification problem where the network estimates the relation between a pixel pair as one of $d_i > d_j$, $d_j > d_i$, or $d_i = d_j$ where $d_i$ and $d_j$ are the depths of the pixel $i$ and pixel $j$, respectively. Since it is infeasible to estimate the relations of all possible pixel pairs given an image, they have utilized the superpixel assumption, and centers of the neighboring superpixels are compared with each other. In order to incorporate long-distance relations, an input image is superpixelated at multiple scales. Once the relative relations are predicted, a global optimization procedure estimates the absolute depth values using predicted relative relations and dataset statistics such as mean and variance of the depth values.
    
    The advantage of this reformulation is twofold. First of all, the pixel-wise regression problem is reduced to a simple three-way classification which allows the usage of well-studied classification methods. Additionally, it is more natural, and hopefully easier to learn to estimate the relative relations of pixels instead of estimating their absolute depth value.
    
        \subsubsection{Datasets}\label{sec:wilddatasets}
        
\begin{figure}[ht]
  \includegraphics[width=\linewidth]{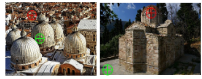}
  \caption{An example input image from DIW dataset. Green points are closer compared to red points. Taken from \cite{chen_single-image_2016}.}
  \label{fig:DIW_example}
\end{figure}

        Chen et al. \cite{chen_single-image_2016} make the first attempt towards depth estimation in the wild and bypass the aforementioned problems by collecting a dataset with relative depth annotations. They crowdsourced the dataset collection process using Amazon Mechanical Turk. Their dataset, called Depth In the Wild (DIW), consists of randomly sampled internet images from a diverse set of scenes. The ground truths consist of the relative relations between two pixels as can be seen in \textbf{Figure \ref{fig:DIW_example},} where green points are annotated as closer by the annotators.

Xian et al. \cite{xian_monocular_2018} introduce another dataset, called ReDWeb. The ReDWeb dataset utilizes stereo images from the internet to reconstruct the corresponding depth maps automatically. Some post processing is applied to the images to improve the quality of the acquired depth maps. This dataset consist of dense relative depth annotations for each image in terms of pixel wise depth scores. The depth score $s_i$ for pixel $i$ and depth score $s_j$ for pixel $j$ are arbitrary numbers that satisfies the relations: 
\begin{equation}
    \label{eqn:depth_score}
    \begin{aligned}
    s_i > s_j, &\text{ if } d_i > d_j\\
    s_i < s_j, &\text{ if } d_i < d_j\\
    s_i = s_j, &\text{ otherwise}.
    \end{aligned}
\end{equation}
        
\begin{figure}[ht]
  \includegraphics[width=\linewidth]{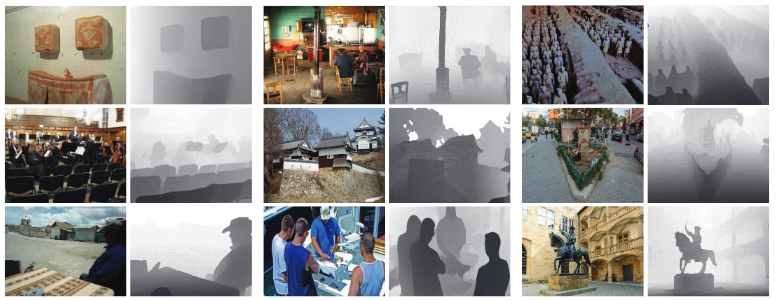}
  \caption{Examples from ReDWeb dataset. Taken from \cite{xian_monocular_2018}.}
  \label{fig:redweb_example}
\end{figure}

Example input ground truth pairs from the dataset can be seen in \textbf{Figure \ref{fig:redweb_example}.}

\begin{figure}[ht]
  \includegraphics[width=\linewidth]{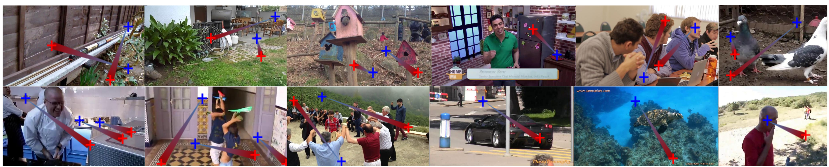}
  \caption{Examples from Youtube3D dataset. Red points are closer than blue points. Taken from \cite{chen_learning_2019}.}
  \label{fig:yt3d_example}
\end{figure}

Lastly, Chen et al. \cite{chen_learning_2019} introduce another dataset called Youtube3D. They use structure from motion (SFM) techniques to automatically label Youtube videos with relative depth annotations. Their ground truths in the dataset consist of relative relations of a varying number of pixels as can be seen in \textbf{Figure \ref{fig:yt3d_example}.}

All the datasets that are designed to work in the wild have relative ground truths. Therefore in this research category, the SIDE problem is reformulated as a relative-SIDE problem, where the aim is to estimate pairwise relations of pixels instead of their absolute depths. Throughout the text, we are going to use the term "relative-SIDE problem" and the term "depth in the wild problem" interchangeably.
        
        \subsubsection{Learning from relative depth annotations}
        
        The main motivation of the aforementioned three works is to present an automatic way to collect and annotate data for depth estimation in the wild problem. Therefore in these papers, the main focus is the dataset. Therefore they use very similar frameworks to tackle the problem.
        
In general, their framework consists of an autoencoder style network that takes the input RGB image and regresses a score map that satisfies the ground truth relative depth annotation in the dataset. The most important part of the framework is the loss function, which uses the ground truth relative depth relations to train the network.

The loss that is used in the work of Chen et al. \cite{chen_single-image_2016} for relative depth estimation is given by
\begin{equation}
    \label{eqn:relative_loss}
    loss=\left\{
    \begin{array}{ll}
    {\log \left(1+\exp \left(-\hat{s}_{i}+\hat{s}_{j}\right)\right),} & {\pi_{ij}=+1} \\ 
    {\log \left(1+\exp \left(\hat{s}_{i}-\hat{s}_{j}\right)\right),} & {\pi_{ij}=-1} \\ 
    {\left(\hat{s}_{i}-\hat{s}_{j}\right)^{2},} & {\pi_{ij}=0}\end{array}\right.,
\end{equation}
where $\hat{s}_{i}$ is the depth score prediction of the network for the pixel $i$, and the $\hat{s}_{j}$ is the depth score prediction for the pixel $j$. $\pi_{ij}$ shows the relative depth relation between the two pixels where $\pi_{ij} = +1$ means pixel $i$ is closer, $\pi_{ij} = -1$ means pixel $j$ is closer and $\pi_{ij} = 0$ means two pixels are at the same depth.

What this loss does is very simple. For $\pi_{ij} = 0$, it penalizes the network for outputting different values. For $\pi_{ij} = +1$ and $\pi_{ij} = -1$, the loss function pushes the network to increase the distance between the estimated depth for the given pixels in the correct direction depending on the relative depth relation \footnote{Note that the actual value that network outputs is irrelevant, only the difference between predicted depth scores affects the loss.}.

Xian et al. \cite{xian_monocular_2018} use the exact same loss with very small modifications. In the previous work, Chen et al. \cite{chen_single-image_2016} experiment with the only available dataset, DIW, which has ground truths of only one pixel pair per image to apply this loss. But in this work, the ReDWeb dataset is introduced that has dense relative depth ground truths in the form of depth score maps. Since it is infeasible to work with all the possible pixel pairs per image, they propose a simple yet effective sampling strategy. They randomly sample 3K pairs of pixels per image, calculate the loss, and use the top 75\% of the pairs with the highest loss. They claim that discarding the low 25\% focuses the learning on the hard pairs. Additionally, there are few pixels that are on the same depth. The proposed loss balances the unequal - equal relations ratio.

Chen et al. \cite{chen_learning_2019} also use the same loss. However, the Youtube3D dataset does not have equality relations in the ground truths, so they disregard the corresponding term.

Xian et al. \cite{xian2020structure} use the same pairwise ranking loss to solve the relative-SIDE problem, but they propose a new sampling strategy that focuses on image and object edges. Instead of randomly sampling all the training points, they sample pixel pairs that reside across instance edges and image edges. While the former one helps the model to gather supervision from object boundaries, which conjectured to be more informative, the latter one helps the model to differentiate between image edges that has a depth discontinuity and image edges that do not have depth discontinuity. 

Instead of improving the sampling strategy, Mertan et al. \cite{mertan2020relative} and Lienen et al. \cite{lienen2020monocular} try to improve the loss function. Both work implements the same idea. They adopt a listwise ranking loss function, the ListMLE, that considers a list of pixels per training instance, as opposed to a pair of pixels. Additionally, \cite{mertan2020relative} advocates the usage of the mean average precision (MAP) as a metric to measure the performance in the test time, instead of accuracy, since MAP is capable of emphasizing the correct ranking of closer pixels, which is arguably more important for the depth estimation problem.

In their later work, Mertan et al. \cite{mertan2020new} focus on another aspect of the solutions, which is the confidence of the estimations. They represent the depth of a pixel as a normal distribution, $d_i \sim \mathcal{N}(\mu_i, \sigma_i^2)$, and try to minimize the following distributional ranking loss (DL) with respect to $d_f$ and $d_c$,
\begin{equation}
    \label{eq:DL}
    \operatorname{DL} = - \log\left( \prod_{f,~c ~\in ~\Omega} P(d_f>d_c)\right)
    ,
\end{equation}
where $\Omega$ represents the set of all pixels, $d_f \sim \mathcal{N}(\mu_f, \sigma_f^2)$ refers to the pixel that is supposed to be farther, and $d_c \sim \mathcal{N}(\mu_c, \sigma_c^2)$ refers to the pixel that is supposed to be closer for a particular pairwise relation. This formulation allows the model to learn per pixel $\sigma$ values that can interpreted as an uncertainty measure for that pixel's estimation.

    \subsection{Other considerably related works}
    In this section, we list some of the works that we encountered along the way that does not fit any category mentioned earlier. Some of them are not even tackling the SIDE problem as we defined. Nonetheless, they are considered interesting work that we want to share without going into too many details.

    \textbf{Depth from Optical Cues} There are number of works that incorporate ideas from optical principles of photography to estimate the depth such as \cite{haim2018depth, wu2019phasecam3d, gur2019single}. These works are especially important as the models learn to estimate the depth from optical cues that are the direct results of physical principles of photography, which are not tied to the semantic structure of the scene, \textit{e.g.} indoor or outdoor, hence are more generalizable. While the neural network models employed in these works are still capable of capturing semantic information and overfitting to the particular regularities of the training set, quantitative cross dataset examination of \cite{gur2019single} strongly indicates that it is not the case. \cite{wu2019phasecam3d} also conjectures that scale ambiguity can be alleviated as well since the optical cues that are learned depend on absolute depth values.
    
    Gur et al. \cite{gur2019single} use defocus cues to solve the SIDE problem. Moreover, they formulate the problem in a self-supervised way where the model learns from supervision signals derived from focused images, instead of ground truth depth information. Their framework consists of two parts. First, a network model estimates the depth of a scene from an all-in-focus image. Next, the estimated depth is used to construct the focused image of the scene in a deterministic and differentiable manner. The reconstruction loss between the focused image generated by estimated depth and the ground truth focused image is used to train the model. Note that this requires a focused image of the scene to be available at training time, which is not the case for mainstream datasets for the SIDE problem. To be able to work on those datasets, they use ground truth depth information to create focused images.

    \textbf{Non-parametric} Karsch et al. \cite{karsch_depth_2014} devise a depth sampling method to estimate the depth without the need for learning any parameter, i.e. non-parametric depth sampling. Their method can be applied to single images as well as videos. For the video case, they use additional steps incorporating temporal relations of frames to increase the quality of the outcome.
    
    The main insight of their method is that similarity in features derived from RGB images means similar depth values. To utilize this insight for depth estimation, they collect a dataset of RGB images and their corresponding depth maps and devised a three-stage method for estimating the depth of a query image. The first stage is to find $k$ candidate images for the query image. Candidate images are found searching the dataset based on their similarity in the RGB domain. In the second stage, warping functions that warps candidate images to the query image are determined using SIFT flow in the RGB domain. In the last stage, depth maps of the candidate images are warped using the warping functions calculated in the second stage, and an optimization procedure is utilized to merge all the depth maps to get the final prediction. 
    
    They were able to achieve relatively good results at that time, without learning any parameters. The important part of their work is they were able to show the validity of their insight which is essentially how statistical methods work as well.  
    
    \textbf{Sensor Fusion} Liao et al. \cite{liao_parse_2017} look at the SIDE problem from a robotics perspective. Even though researchers are getting better and better results metrically, the scale ambiguity is still a problem, and the success of the performance that is measured with the available metrics is not necessarily a good indication of success, particularly when we deploy estimated depths for a robotics task.
    
    To increase the performance and the applicability of the acquired solutions to the robotics tasks, additional sensory information is incorporated as an input to their system. To this end, they experiment with a 2D laser range finder, which is cheap and already available in lots of robots.
    
    As an input, the network of Liao et al. takes an RGB image and a reference depth map which is created by extending the 2D laser scan values vertically to the ground plane. By doing that, a 2D planar depth observation can be turned into a dense 2D reference depth map by projecting it along the gravity direction. The reference depth map can be concatenated to the input image and together they are fed to the network. The network then does a pixel-wise classification to estimate a residual depth map, showing the difference between the reference depth map and the ground truth depth map. Moreover, they utilize a regression loss combined with the classification loss, which allows them to make use of ordinal relations between classes similar to the idea of \cite{fu_deep_2018}.
    
    The main contribution of this work is the usage of a laser scan, which helps alleviate the scale ambiguity problem, resulting in better solutions. To the best of our knowledge, they were the first to do "depth completion", a new branch of research where the aim is to utilize sparse ground truth depth information alongside the RGB image as input, mimicking additional sensory information.
    
    Mal and Karaman \cite{mal_sparse--dense:_2018} also work on depth completion. Their framework is very similar to previous works, with the single contribution of an additional input they choose to use. Instead of using 2D laser scan information, they mimic structured light-based sensors and use the depth information of randomly sampled points across the image domain as an additional input. This can also be viewed as super-resolution for sensors, which is demonstrated as an application in their work.
    
    \textbf{3D Reconstruction} Izadinia et al. \cite{izadinia_im2cad_2017} design a system that infers the 3D CAD model of a given scene, inspired by the 1963 Ph.D. thesis of Lawrence Roberts \cite{roberts_machine_1963}. Using a database of known objects, Roberts's system was able to infer the 3D scene from a given single RGB image, even with back-facing objects or occlusions. Even though a lot of progress has been made in computer vision since then, it is still a very challenging problem to estimate a 3D model of a complex, natural scene.
    
    To tackle the problem, Izadiana et al. \cite{izadinia_im2cad_2017} develop a pipeline that processes the input RGB image and estimates the 3D CAD model of the scene using an object database containing CAD models of known objects. Their pipeline consists of roughly two parallel processes. 
    
    In one process, an off the shelf object detector is used to detect objects in the given scene. Using the CAD models of the detected objects, possible alignments of them are rendered and compared to the input RGB image to infer their alignment in the scene.
    
    In the second process, pixels are semantically labeled as floor, wall, or ceiling in order to be able to estimate the layout of the scene. Essentially, a box is fitted with the assumption of indoor scenes.
    
    Having the room layout and objects with their alignments, the 3D scene is reconstructed. As the last step, the reconstructed scene is used as an initial estimate, and it is further refined through optimization by rendering and comparing it to the input image.
    
    Even though there are obvious fail cases, such as object detector missing the object, object database not containing an entry for the detected object, or room layout not being box-shaped, they were able to show that inferring qualitatively pleasing 3D scenes from a natural, complex single image is possible.

\section{Investigative Works}

Dijk and Croon \cite{dijk_how_2019} investigate the networks of Godard et al. \cite{godard_unsupervised_2017}, Zhou et al. \cite{zhou_unsupervised_2017}, Kuznietsov et al. \cite{kuznietsov_semi-supervised_2017}, and Wang et al. \cite{wang_learning_2018} in an autonomous driving scenario. They make their analysis using probe images designed by hand to reveal high-level behaviours of the networks, instead of examining feature map visualizations which can only give insights about the low-level understanding of the networks.

First, they focus on revealing how objects' positions in the image and their apparent sizes affect the depth estimation. To this end, they crop images of cars and insert them into other images in vertically different positions keeping the apparent size the same, same position with different apparent sizes, and vertically different positions with an appropriate size change. Networks' depth estimation for the inserted car is reduced to a scalar by averaging the estimated depth on a flat surface of the front or rear end of the car, depending on whether the inserted car faces the camera or not. They show that the networks mostly use vertical position cue when estimating the depth, disregarding the apparent size cue. Moreover, this behaviour is shown to be consistent in all the investigated networks which have different architectures and training methods.

They also experiment on Godard et al.'s network \cite{godard_unsupervised_2017} to see how different camera setups affect the estimated depth of objects. Cropping images with different vertical offsets, they mimic the pitch change of the camera and they show that the pitch changes affect the estimated depth values, which further strengthen the claim of the vertical position being the most influential cue.

As another experiment, they investigate whether the inserted object's properties such as color, shape and so on, are important. They show that the network essentially tries to detect a connection point to the ground to estimate the depth of objects, in fact when such a connection is not detected, the network ignores the object altogether. Furthermore, even when the object's inside is emptied, the network still fills in the interior in the estimated depth, showing that edges are guiding the depth estimation process. While grayscale objects or falsely colored objects do not degrade the performance too much, objects that are colored completely to a single color degrades the performance showing that the important thing for the network is to be able to distinguish parts of the objects, not their actual color or texture.

Hu et al. \cite{hu_visualization_2019} also attempt to find an answer to the questions of how neural nets estimate the depth from a single image; which cues do they use? Instead of visualizing intermediate feature maps, they also work on the input image. Unlike \cite{dijk_how_2019}, their goal is to see which parts of the input image have a bigger impact on the estimated depth. They formulate this as an optimization problem where a network learns to mask out irrelevant pixels from the input image that results in a minimal change in the estimated depth map from the entire image.

To clarify, first they train a network $N$ for the SIDE task on the NYU and KITTI datasets with common architectures \footnote{All experimented architectures result in similar findings showing that it is not dependent on any particular architecture, therefore we are not going to dive into details of them.} seen in the literature such as ResNet50 in Laina et al. \cite{laina_deeper_2016}. Afterward, while the weights of the network $N$ is fixed, a mask predicting network $G$ is trained to optimize the following problem:
\begin{equation}
    \label{eqn:visualizationmask}
    \min _{G} ~~l_{\mathrm{dif}}~(~N(I) ,~ N(I \otimes G(I)))+\lambda \frac{1}{n}\|G(I)\|_{1},
\end{equation}
where $\|G(I)\|_1$ pushes the network to mask out more pixels (otherwise a trivial solution to the optimization problem would be just outputting all 1s), $\lambda$ controls its weight, $n$ is the number of pixels, and  $l_\mathrm{dif}$ is the function calculating the difference between estimated depth maps that consists of a depth loss, a depth gradient loss, and a depth normal loss. Here, $\otimes$ refers to an elementwise multiplication.

When the predicted masks are examined, they are able to show that the edges are utilized by the networks to estimate the depth. However, not all image edges are present in the masks, which claimed to be the indication of the network's ability to distinguish informative edges. Another important finding is that the interiors of some of the objects look important to the network. This looks especially true for smaller objects. This finding may seem to be contradicting with the results of Dijk and Croon \cite{dijk_how_2019} at first. However, the latter work only experimented with cars, and it is possible that the networks that are examined may have overfitted to cars. One last important finding is the importance of the vanishing points for outdoor images. Network models seem to be paying great attention to those areas of images. They conjecture that this might very well be the result of errors in those pixels being larger due to their depth, however, they may also be more important for understanding the geometry of the scene.

\subsection{Discussion}
The investigative works show that models learn to exploit regularities in their training set. Simple cues such as ground plane connection, vanishing points, and object edges are utilized to estimate depth. However, it is questionable how well these cues generalize outside of their training domain. We believe it is important to systematically investigate how well pictorial cues are utilized by the models and how the learning can be structured in a way that models learn these types of generalizable cues. In addition to that, investigations show that datasets cause biases and affect the learned cues. In order to be able to have a model that can work in the wild, it is crucial to have diverse datasets. In that sense, these investigative works justify in the wild datasets that we discuss in \textbf{ Section \ref{sec:wilddatasets}. }

In addition to the failure cases shown in \cite{dijk_how_2019}, qualitative results of number of works such as \cite{mertan2020new} reveals that models fail to estimate the depth correctly for surfaces with figures or reflective surfaces such as mirrors. Intuitively, estimating the depth correctly in these situations requires high-level reasoning, e.g. model needs to correctly identify the mirror and ignore the features for that region in the input image to estimate a plain depth.

\section{Conclusion}

In this work, we investigate the SIDE problem by providing an overview of existing solutions, identifying the main research directions, and emphasizing common themes among works. Additionally, we discuss investigative works that give insight on how the monocular depth estimation is done by the models. This work is by no means an exhaustive survey. Our aim is to simply give an overview of the field.

Based on our work, we list some of the research directions and open questions, as well as specific future work suggestions for the SIDE problem.

\textbf{Absolute depth in the wild} While there are works such as \cite{gur2019single} that show promising results for cross dataset examination, we believe that it is still an important research direction that requires more attention. A more general dataset is needed to train and test the models. 

\textbf{Increasing the robustness} Seemingly trivial things such as lack of ground connection, surfaces with figures or reflective surfaces such as mirrors can easily disrupt the estimation of the models. An in depth investigation of such failure cases and methods to increase the robustness of the solutions are necessary for safe deployment of these solutions in real world applications.

\textbf{Ranking losses} Losses we have discussed for the relative-SIDE problem are called surrogate losses, as we are not directly optimizing ranking measures such as MAP and NDCG due to their not being differentiable. In recent years, relaxations of ranking metrics that make them differentiable have been devised, allowing them to be used in the optimization process directly \cite{cakir_deep_2019, revaud_learning_2019, grover_stochastic_2019}. These new losses should be investigated in the context of the relative-SIDE problem. Furthermore, \cite{cuturi_differentiable_2019, mena_sinkhorn_2017} proposed new approaches for ranking based on the optimal transport. Their potential success in high-level tasks such as the relative-SIDE can be examined.

\textbf{Analysis of alternative solutions} So far, models that estimate the absolute depth, such as \cite{godard_unsupervised_2017, zhou_unsupervised_2017, kuznietsov_semi-supervised_2017, wang_learning_2018}, are investigated. Models that are trained to work in the wild by estimating relative depth and models trained to learn optical cues should be investigated as well, as their methodology differs from the previously examined models.

\textbf{Absolute depth from relative depth} Similar to the idea of using additional sparse sensory information in depth completion in \cite{liao_parse_2017, mal_sparse--dense:_2018}, one can use off-the-shelf relative depth estimator models as they are designed to work in the wild, and estimate absolute depth with additional relative depth information alongside RGB image. The success of this approach in terms of metric performance, the need for labeled data, and the ability to adapt to cross-domain scenarios can be investigated.       

\textbf{3D structure} Using camera parameters, the depth map of a scene can be projected onto 3D. So far, researchers only qualitatively analyzed 3D projected depth maps. It has been claimed that the methods may exhibit qualitatively different performances even though they have very similar quantitative performances in the metrics calculated from depth maps \cite{li_two-streamed_2017}. A systematic way to measure the quantitative performance of 3D projections of depth maps needs to be devised. 

\textbf{Embodied AI} We expect the SIDE models to learn pictorial cues that are used by the human visual system to estimate depth, yet our models are trained very differently when compared to how humans learn. While humans learn and build representations of the world actively through interactions with the world, models that we train learn passively from a dataset of static examples. It is possible that agents that learn  through interactions may build better representations of the world for the tasks they are optimized to solve. Similar to \cite{pinto_curious_2016}, agents that learn to predict depth through interacting with the world can be an interesting future research direction.

\section*{Acknowledgments}

This work is supported by the Scientific and Technological Research Council of Turkey (TÜBITAK), project 116E167.

\newpage
\bibliographystyle{IEEEtran}
\bibliography{depthEstimation.bib}

\end{document}